\documentclass[sigconf]{acmart}

\AtBeginDocument{%
  }

\usepackage[nolist]{acronym}
\usepackage{makecell}
\pretocmd{\section}{\acresetall}{}{}

\usepackage{parskip}
\usepackage{graphicx}
\usepackage{natbib}

\setcopyright{none} 
\renewcommand\footnotetextcopyrightpermission[1]{} 

\begin{document}

\title{SatBLIP: Context Understanding and Feature Identification from Satellite Imagery with Vision–Language Learning}

\author{Xue Wu}
\affiliation{%
  \institution{The University of Alabama}
  \city{Tuscaloosa}
  \state{Alabama}
  \country{USA}
}
\email{xwu33@crimson.ua.edu}
\orcid{0000-0002-0048-9008}

\author{Shengting Cao}
\affiliation{%
  \institution{Knox College}
  \city{Galesburg}
  \state{Illinois}
  \country{USA}
}
\email{scao@knox.edu}
\orcid{0000-0001-6175-0831}

\author{Shenglin Li}
\affiliation{%
  \institution{The University of Alabama}
  \city{Tuscaloosa}
  \state{Alabama}
  \country{USA}
}
\email{sli90@ua.edu}
\orcid{0009-0002-3608-8369}

\author{Jiaqi Gong}
\affiliation{%
  \institution{The University of Alabama}
  \city{Tuscaloosa}
  \state{Alabama}
  \country{USA}
}
\email{jiaqi.gong@ua.edu}
\orcid{0000-0001-9694-2518}




\begin{abstract}
Rural environmental risks are shaped by place-based conditions (e.g., housing quality, road access, land-surface patterns), yet standard vulnerability indices are coarse and provide limited insight into risk contexts. We propose SatBLIP, a satellite-specific vision–language framework for rural context understanding and feature identification that predicts county-level Social Vulnerability Index (SVI). SatBLIP addresses limitations of prior remote-sensing pipelines—handcrafted features, manual virtual audits, and natural-image–trained VLMs—by coupling contrastive image–text alignment with bootstrapped captioning tailored to satellite semantics. We use GPT-4o to generate structured descriptions of satellite tiles (roof type/condition, house size, yard attributes, greenery, and road context), then fine-tune a satellite-adapted BLIP model to generate captions for unseen images. Captions are encoded with CLIP and fused with LLM-derived embeddings via attention for SVI estimation under spatial aggregation. Using SHAP, we identify salient attributes (e.g., roof form/condition, street width, vegetation, cars/open space) that consistently drive robust predictions, enabling interpretable mapping of rural risk environments.
\end{abstract}






\maketitle
\pagestyle{empty}

\section{Introduction}
Satellite imagery offers a rich data source for extracting features of rural environments and understanding their socioeconomic and health implications \cite{Wu2022}. In recent years, geographic information derived from satellite images has become increasingly important for modeling tasks across diverse fields such as ecology and epidemiology \cite{Klemmer2025, Wu2025, Wu2026}. These images capture both natural and built-environment characteristics, providing fine-grained spatial information essential for environmental and public health research \cite{Roberts}.

Traditional satellite analysis often rely on multilevel socioecological frameworks with handcrafted features\cite{Watmough2019}, systematic virtual audits using Google Earth with manual annotation\cite{Crawford2019}, or object detection techniques\cite{Ayush2020}. However, these methods may not generalize well to the diverse and heterogeneous characteristics of rural environments. Additionally, many existing vision-language models CLIP  and BLIP\cite{Radford2021, li2022blip} are trained on natural images paired with curated captions, which limits their effectiveness when applied to satellite imagery that involves domain-specific and context-dependent semantics. Furthermore, current methods\cite{Crawford2019, Ni2020} often fail to extract interpretable, high-level semantic representations—an essential capability for modeling complex social vulnerability factors in rural settings.

With the rapid advancement of multimodal large language models (MLLMs), new opportunities are emerging to explore geospatial and geographic domains through joint vision-language modeling\cite{Roberts}. Large language models (LLMs), in particular, have demonstrated remarkable capabilities in interpreting complex semantic information, enabling a broader understanding of visual content across various domains.

Integrating vision and language learning from satellite data presents a powerful approach for analyzing rural environments—such as identifying regional features, monitoring ecological conditions, and estimating socioeconomic indicators\cite{Roberts}. By leveraging LLMs, it becomes possible to extract and reason about fine-grained visual details in satellite 
imagery\cite{Wang2023, Singh}. This joint representation facilitates more interpretable, scalable, and effective modeling of rural risk environments and supports progress in environmental, public health, and geographic research.

Leveraging this potential, recent vision–language models offer a path beyond traditional limitations. We propose a satellite-based Vision–Language Pretraining (VLP) framework that integrates contrastive learning with large language models to understand rural environments from satellite imagery. Building on BLIP, SatBLIP bootstraps captions from noisy data to generate detailed, context-aware descriptions, which are used to extract fine-grained semantic features for interpretable and accurate prediction of the Social Vulnerability Index (SVI).

In this study, we introduce a satellite-based vision-language framework that leverages contrastive learning and bootstrapped captioning to generate fine-grained, semantically meaningful descriptions of satellite imagery. Our proposed pipeline that transforms textual features from satellite captions into predictive signals for county-level SVI, and further identifies satellite features that are salient to SVI.

\section{Related Work}
Large language models (LLMs) have significantly accelerated progress toward general-purpose artificial intelligence, demonstrating strong zero-shot and few-shot performance across a wide range of tasks \cite{Wang2023}. Building on this foundation, VisionLLM \cite{Wang2023} introduces a unified framework for multimodal task execution using language instructions, enabling both fine-grained object-level and coarse-grained task-level customization. Recent LLMs, such as ChatGPT and GPT-4, have shown promising capabilities in satellite image interpretation and environmental reasoning \cite{Chen2025}.

Several recent works have explored the integration of satellite imagery and language models for geographic and health-related applications. For instance, Chen et al. \cite{Chen2025} proposed a simplicial contrastive learning framework leveraging LLMs to uncover latent urban features and their associations with air pollution (PM2.5). \cite{Moukheiber2024} developed a reproducible multimodal pipeline for satellite image embedding generation, enabling enhanced decision support in public health contexts. \cite{Dhakal} introduced Sat2Cap, a vision-language framework that maps geolocations to fine-grained textual prompts, producing high-quality zero-shot semantic maps.

In the remote sensing domain, \cite{Klemmer2025} proposed SatCLIP, which aligns satellite imagery with geographic coordinates by contrasting features from convolutional neural networks and vision transformers. \cite{Jean} combined satellite imagery with machine learning to predict poverty from high-resolution images in sub-Saharan Africa. \cite{Wu2022} designed a cost-effective sampling strategy using satellite imagery to study COVID-19 vaccine uptake in rural Alabama. \cite{Roberts} explored the geographic capabilities of frontier vision-language models like GPT-4V, while \cite{Crawford2019} performed virtual audits using Google Earth to identify built environment features in rural Kentucky counties.

While these studies demonstrate the promise of combining satellite imagery and language models, most existing approaches either focus on urban settings, require curated prompts, or lack generalizability to rural and underserved communities. Moreover, few works systematically connect vision-language representations to social vulnerability indicators in a fully self-supervised manner.


\section{Methods}




\subsection{Zero-Shot Prompts for Understanding Satellite Imagery}
We integrate large language models (LLMs) into satellite imagery analysis to construct a vision-language representation framework aimed at extracting meaningful housing and environmental features. To structure the representation task, we design a 5-tier prompt framework, each tier targeting a specific level of detail or abstraction from the satellite images:

\begin{itemize}
\item Tier 1 – Visual Summary Prompting: We begin with general prompts to generate a natural language summary of the satellite image. This functions similarly to traditional computer vision object detection, providing an overview of prominent features such as buildings, roads, or vegetation.

\item Tier 2 – Objective Feature Prompting: This tier introduces structured prompts focused on clearly measurable housing characteristics, such as roof type, building size, or lot shape.

\item Tier 3 – Neighborhood and Environmental Context: We design prompts to extract contextual information about the surrounding environment, including yard presence, tree coverage, nearby infrastructure, and proximity to roads or other buildings.

\item Tier 4 – Geographic and Socioeconomic Reasoning: Leveraging the background knowledge embedded in LLMs, we use prompts that explore higher-level geographic and socioeconomic associations, such as indicators of wealth, urbanization, or vulnerability.

\item Tier 5 – Interpretive Summary Generation: In the final tier, we prompt the model to generate concise interpretive summaries that combine visual evidence with learned reasoning, producing rich, human-readable descriptions.
\end{itemize}

\begin{figure}[!t]
\centering
\includegraphics[scale=.25]{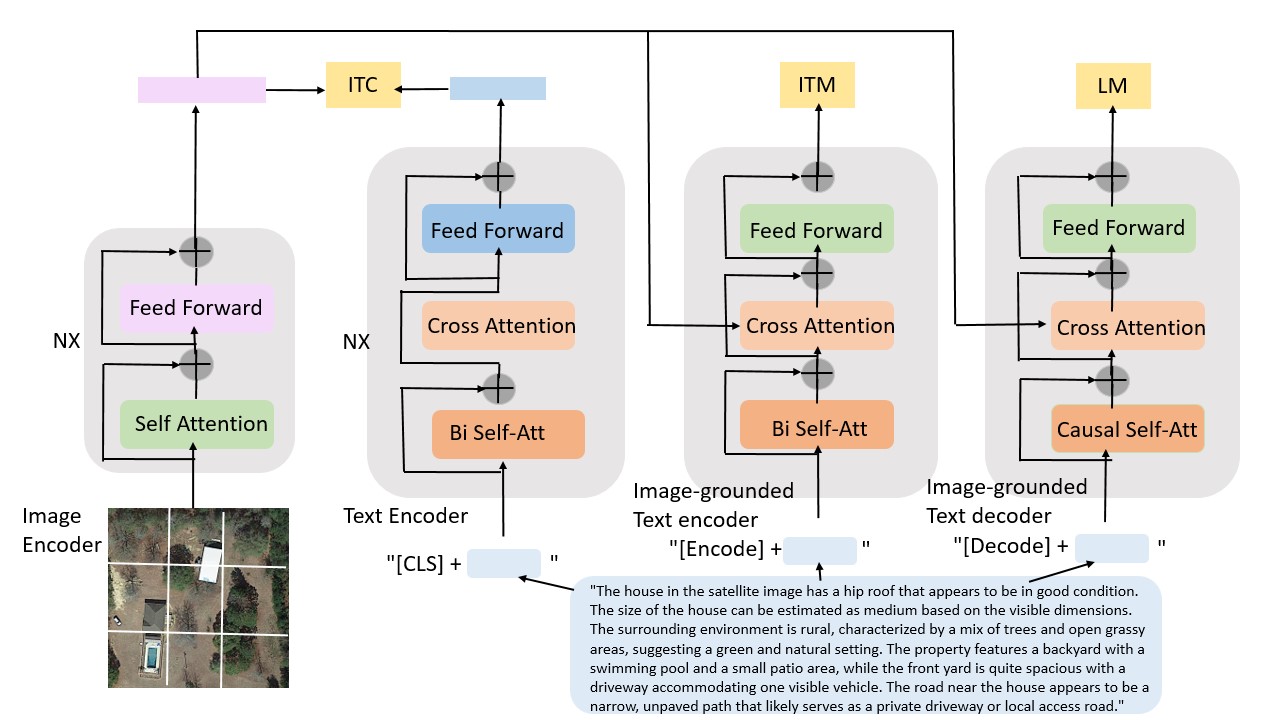}
\caption{Satellite based BLIP model}
\label{satblip}
\end{figure}

For our analysis, we train a satellite-specific version of the BLIP (Bootstrapped Language-Image Pretraining) model to generate these descriptions and support structured representation learning across tiers. As shown in Figure \ref{satblip}, this hierarchical framework employs a vision transformer to encode images and a text encoder to process text, with image–text similarity learned via image–text contrastive (ITC) loss. The image-grounded text encoder integrates both modalities, and parameters are shared across blocks of the same color.

\subsection{Zero-Shot Captioning and Multimodal Embedding with LLMs}
We utilize large language models (LLMs) to enhance the understanding of satellite imagery through zero-shot image-to-text generation. Specifically, we define a structured prompt below to guide the model’s attention toward relevant visual attributes:

\begin{quote}
\textit{``Analyze the satellite image and provide a detailed description of: House roof type and condition (e.g., metal, shingles, new, old, damaged); House size (small, medium, large, estimated square footage); Surrounding environment (urban, rural, trees, greenery, desert, snow, etc.); Front yard and backyard details (size, garden, pool, patio, driveway); Nearby road situation (wide road, narrow street, highway, gravel road).''}
\end{quote}

This prompt is passed to the LLM along with the satellite image input $I$, enabling it to generate rich, descriptive captions that capture multiple dimensions of the built and natural environment. Once the description is generated, we apply a vision-language model (VLM) to extract semantic embeddings from the image-text pair. These embeddings are then passed through a multi-layer perceptron (MLP) to produce a refined representation of the satellite image. The final representation is defined as:

\begin{equation}
\Xi_{\text{LLM}} = f_{\text{MLP}} \left( \mathrm{VLM} \left( \mathrm{LLM} \left( \text{Prompt}(I) \right) \right) \right)
\end{equation}

where $f_{\text{MLP}}$ is a two-layer perceptron with batch normalization. To align with GPU memory constraints and accelerate matrix operations, we utilize CLIP (ViT-B/32) to encode satellite image descriptions into 512-dimensional embeddings.

Since the semantic relevance of different modalities may vary, we introduce an attention mechanism to dynamically weight the contributions of each embedding for downstream tasks. Specifically, we compute attention-weighted fusion between the satellite-caption embedding $\Xi_{\text{SC}k}$ and the LLM-derived embedding $\Xi_{\text{LLM}}$:

\begin{equation}
(\rho_1, \rho_2) = \text{Att}(\Xi_{\text{SC}k}, \Xi_{\text{LLM}})
\end{equation}

\begin{equation}
\rho_i = \text{softmax}_i \left( \Upsilon_{\text{Att}} \tanh \left( \Phi \Xi_i \right) \right)
\end{equation}

where $\Upsilon_{\text{Att}} \in \mathbb{R}^{1 \times d_O}$ is a trainable linear transformation, $\Phi$ is a learnable weight matrix, and $d_O$ is the output dimension. The softmax ensures that the attention coefficients across modalities sum to 1. Finally, the fused multimodal embedding is computed as:

\begin{equation}
\Xi_O = \rho_{\text{SC}k} \cdot \Xi_{\text{SC}k} + \rho_{\text{LLM}} \cdot \Xi_{\text{LLM}}, \quad \Xi_{\text{SC}k}, \Xi_{\text{LLM}}, \Xi_O \in \mathbb{R}^{512}
\end{equation}

This fused representation $\Xi_O$ enables robust multimodal reasoning and supports interpretable downstream prediction tasks such as Social Vulnerability Index (SVI) estimation.

\subsection{Fine-Grained Feature Descriptions Model for Satellite Images}
After getting synthetic satellite description using OpenAI’s GPT-4o API. We train a satellite based BLIP model using the paired satellite images and satellite descriptions. This proposed satellite-based vision-language framework that integrates the BLIP (Bootstrapped Language-Image Pretraining) model \cite{li2022blip} to enhance semantic understanding and mapping of satellite imagery. We fine-tune BLIP using our collected dataset of satellite images paired with natural language descriptions generated by the GPT-4o model. Through this process, the resulting model becomes specialized in generating rich and context-specific descriptions for satellite images similar to those in the training set. 

\begin{figure*}[t] \setlength\belowcaptionskip{-1.5\baselineskip}\setlength\abovecaptionskip{-0.2\baselineskip}
  \centering
  \includegraphics[width=\textwidth]{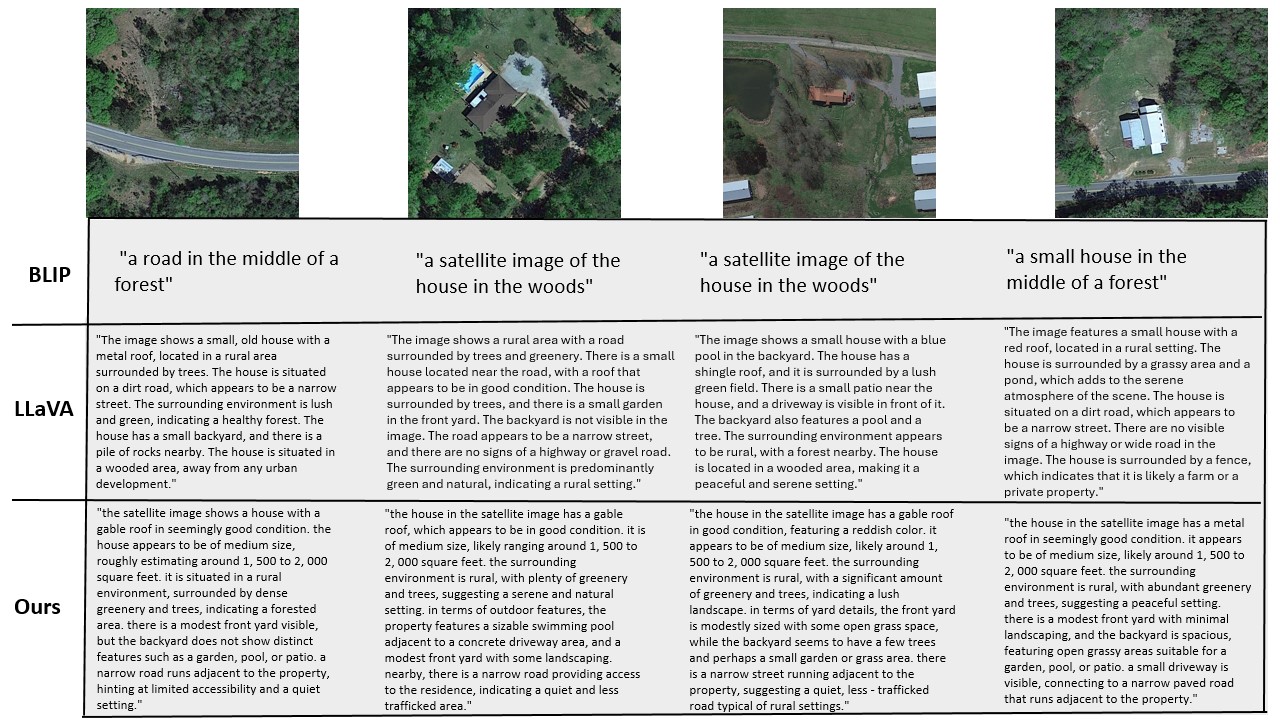}
  
  \vspace{0.5em} \caption{Descriptions generated using BLIP model for satellite images with Pre-trained checkpoints, LLaVA, and our trained satellite-based BLIP model.}
  \vspace{10pt}
  \label{blip_satellite_des}
\end{figure*}

This approach enables flexible alignment between satellite images and free-form textual descriptions, facilitating zero-shot generalization across a variety of vision-language tasks in the remote sensing domain. Specifically, the fine-tuned BLIP model is employed to produce detailed, interpretable, and context-aware descriptions for previously unseen satellite images, improving semantic grounding and enhancing the interpretability of remote sensing analyses. Figure \ref{blip_satellite_des} presents satellite image descriptions generated by the BLIP model with pre-trained checkpoints, LLaVA, and our fine-tuned satellite-based BLIP model. Among these, our satellite-trained BLIP model produces descriptions that more accurately reflect the visual content, offering greater detail and precision. It captures specific features while avoiding factual inaccuracies (e.g., LLaVA described a paved road as dirt) and unsupported negative assumptions (e.g., LLaVA labeled a house as “old” without visual evidence). Furthermore, it adopts a more informative tone, which is better suited for downstream analytical tasks.

This work contributes to the field of multimodal AI for satellite image analysis by leveraging pre-trained vision-language models to learn and generalize complex image-text relationships. While prior research has explored cross-modal alignment in remote sensing, including methods for learning joint visual-textual representations from satellite imagery \cite{Dhakal}, our framework builds upon well-established methods by demonstrating strong zero-shot generalization to previously unseen satellite images. This capability enables flexible and scalable interpretation of remote sensing data through natural language, unlocking new avenues for automated geospatial intelligence and significantly expanding the potential of language-guided applications in Earth observation and environmental monitoring.

In addition to analyzing physical features, we examine the textual information derived from satellite imagery to further investigate the Social Vulnerability Index (SVI) in rural areas \cite{Flanagan2020}. Our objective is to develop and validate a specialized text-processing pipeline tailored for satellite-based analysis within rural risk environments \cite{Rhodes2002, Rhodes2009}. The natural language descriptions generated from our satellite dataset capture critical attributes, such as roof condition, yard presence and quality, surrounding environmental context, and nearby road infrastructure and traffic conditions. These textual features exhibit strong alignment with several components of the SVI, particularly the Household Composition theme and the Housing Type \& Transportation theme, enabling a novel form of semantic mapping between satellite observations and social vulnerability indicators.

\begin{figure}[!t]
\centering
\includegraphics[scale=.224]{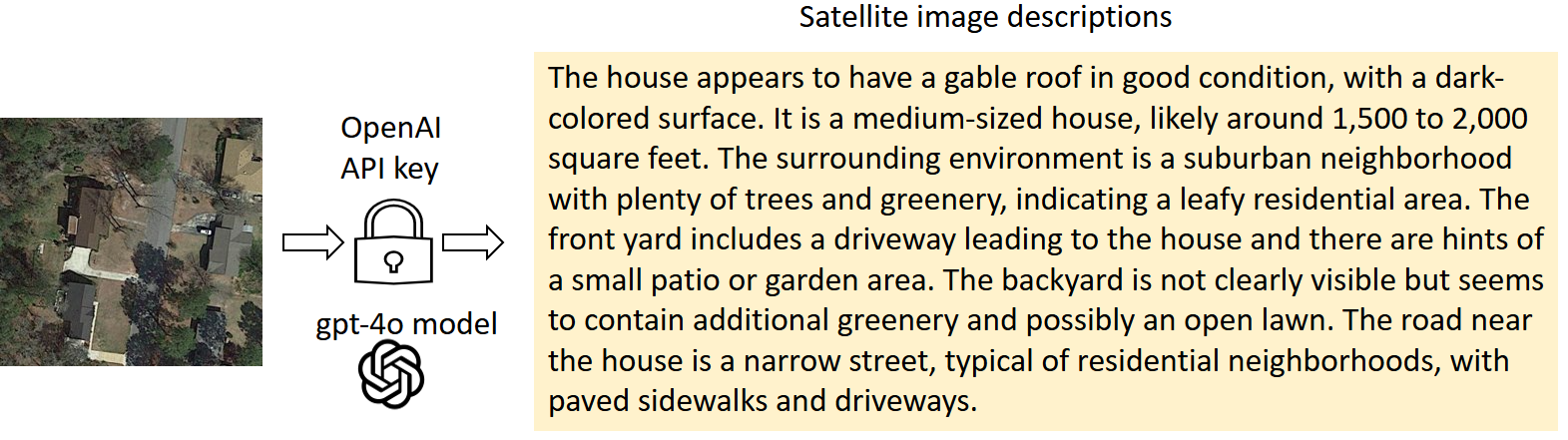}
\caption{We use OpenAI's GPT-4o model to generate synthetic descriptions for satellite images.}
\label{satdes_gpt}
\end{figure}

\begin{figure*}[t] \setlength\belowcaptionskip{-1.5\baselineskip}\setlength\abovecaptionskip{-0.2\baselineskip}
  \centering
  \includegraphics[width=\textwidth]{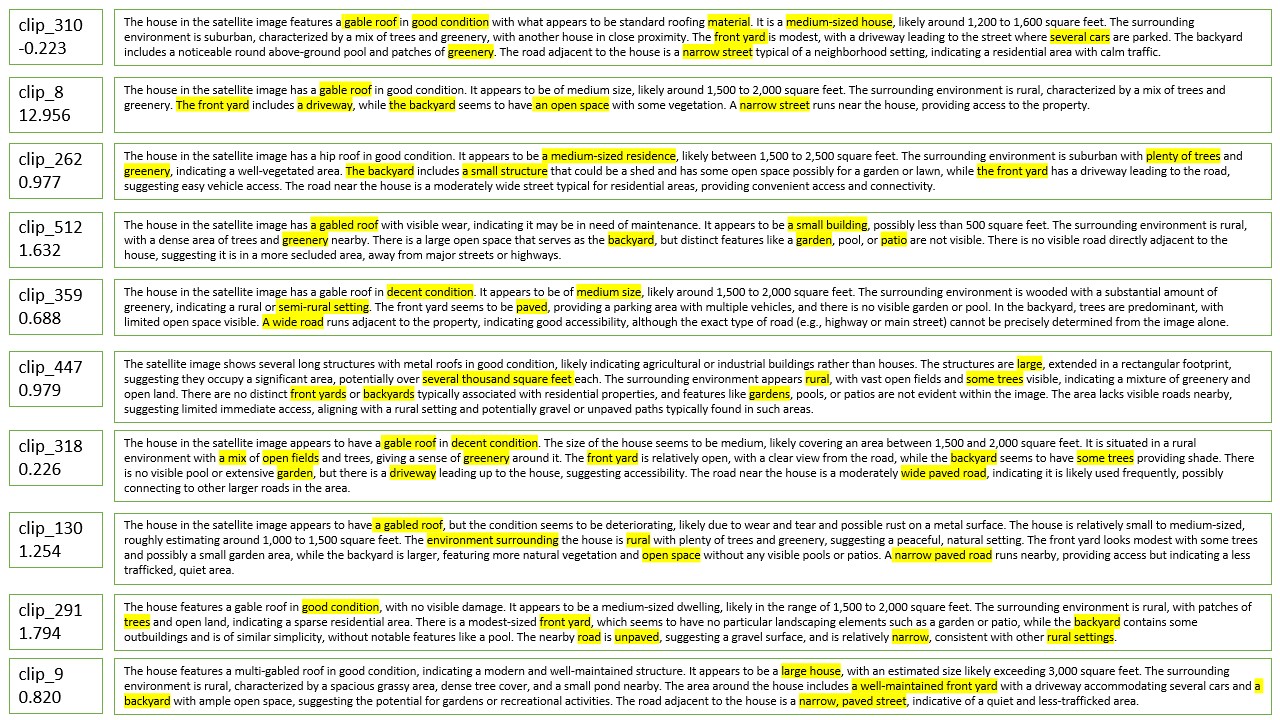}
  \caption{Top 10 SHAP-important dimensions with top description per dimention, highlighted with most frequenctly appeared  features among top 5 descriptions per dimension}
  \label{embedding_description}
\end{figure*}

\section{Result}
\subsection{Explanations via SHAP}
We explain the impact of individual embedding dimensions on SVI (Social Vulnerability Index) predictions using SHAP (SHapley Additive exPlanations) \cite{Lundberg}. SHAP is a game-theoretic approach that provides feature-level interpretability for machine learning models by quantifying each feature’s contribution to a given prediction.

To identify the most influential dimensions in our model, we compute and visualize SHAP values for the top 10 CLIP embedding dimensions used in predicting SVI from satellite image descriptions. Each embedding dimension represents a latent feature learned by the CLIP text encoder, corresponding to an abstract direction in high-dimensional semantic space. While not directly interpretable, these dimensions capture meaningful patterns such as building density, vegetation, infrastructure, or land use context. Additionally, we highlight the five satellite descriptions with the highest SHAP importance values, illustrating how specific textual features contribute most significantly to the model’s prediction of social vulnerability.

\begin{figure}[!t]
\centering
\includegraphics[scale=.35]{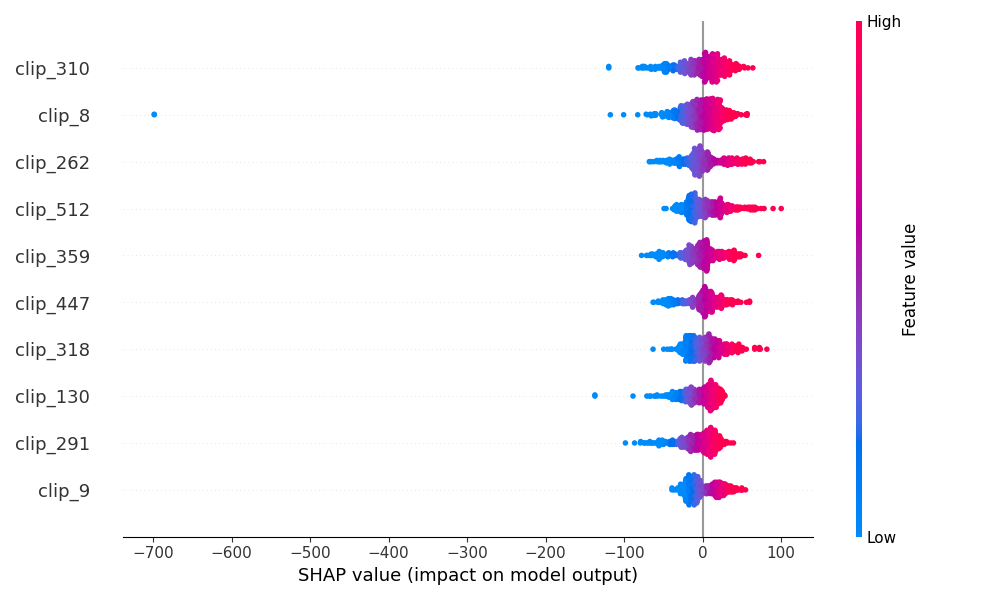}
\caption{Summary of the effects of top 10 satellite description embedding dimensions.}
\label{embedding_svi_SHAP}
\end{figure}

\subsection{Feature representation}
To interpret the latent dimensions identified as most important by SHAP, we extract the top five satellite image descriptions associated with each of the top 10 SHAP-ranked embedding dimensions. By analyzing the most frequently occurring semantic features within these descriptions, we infer the potential meaning of each dimension in relation to social vulnerability. In Figure~\ref{embedding_description}, we highlight the semantic interpretations of these influential embedding dimensions and their impact on SVI predictions. This analysis provides insight into how abstract features captured by the CLIP model—such as housing density, road conditions, or environmental context—relate to patterns of social vulnerability in rural areas. Overall, our research demonstrates how combining satellite imagery with large language models enables the recognition of meaningful features in rural environments and facilitates the exploration of their associations with environmental and socioeconomic variations.


\section{Conclusion}
In this paper, we identify rural features associated with the Social Vulnerability Index (SVI) through a multimodal integration of large language models and satellite imagery using contrastive learning. We evaluate our approach using multiple regression models to assess the predictive power of vision-language representations. Our results show that county-level prediction outperforms image-level prediction, as it better mitigates noise from visually similar images across counties. We also find that certain textual features extracted from satellite image descriptions—such as gable roof, good condition, medium-size house, narrow street, greenery, several cars, open space, and rural settings—contribute substantially to SVI prediction. These interpretable features help reveal environmental patterns associated with social vulnerability in rural areas. In future work, we plan to explore the integration of multiple language models and multi-year satellite imagery to study temporal dynamics and further improve the understanding of rural risk environments.

\bibliographystyle{ACM-Reference-Format}
\bibliography{main}
\end{document}